\documentclass[acmsmall,screen]{acmart}

\AtBeginDocument{%
  }

\setcopyright{acmlicensed}
\copyrightyear{2024}
\acmYear{2024}
\acmDOI{XXXXXXX.XXXXXXX}

\acmJournal{TOMM}
\acmVolume{0}
\acmNumber{0}
\acmArticle{0}
\acmMonth{0}

\usepackage{hyperref}
\usepackage{booktabs}
\usepackage{pgfplots}
\usepackage{float}
\usepackage{graphicx} 
\usepackage{subcaption}
\usepackage{tikz}
\usepackage{url}
\usepackage{multirow}
\usepackage{csquotes}
\usepackage{pifont}

\usepackage{xspace}
\def\scns{\textsc{Scn}\xspace}
\def\schem{\textsc{SocialChemistry101}\xspace}
\def\normb{\textsc{NormBank}\xspace}
\def\norms{\textsc{NormSage}\xspace}
\def\prompt{\textsc{Prompt}\xspace}
\def\sdial{\textsc{SocialDial}\xspace}
\def\cnormb{\textsc{ChineseNormBase}\xspace}
\def\ndial{\textsc{NormDial}\xspace}

\usepackage{ulem}

\usepackage{lipsum}
\usepackage{enumitem}

\usepackage{amsthm,amsmath,amsfonts,bm,xspace}
\usepackage{color}

\definecolor{mgreen}{rgb}{0,0.7,0}

\begin{document}


\title{Scalable Frame-based Construction of Sociocultural NormBases for Socially-Aware Dialogues}
\author{Shilin Qu}
\email{shilin.qu@monash.edu}
\author{Weiqing Wang}
\email{teresa.wang@monash.edu}
\author{Xin Zhou}
\email{xin.zhou@monash.edu}
\author{Haolan Zhan}
\email{haolan.zhan@monash.edu}
\author{Zhuang Li}
\email{zhuang.li@monash.edu}
\author{Lizhen Qu}
\email{lizhen.qu@monash.edu}
\author{Linhao Luo}
\email{linhao.luo@monash.edu}
\author{Yuan-Fang Li}
\email{yuanfang.li@monash.edu}
\author[1]{Gholamreza Haffari}
\email{gholamreza.haffari@monash.edu}
\affiliation{%
  \institution{\\Faculty of Information Technology, Monash University}
  \city{Melbourne}
  \country{Australia}
}




\renewcommand{\shortauthors}{Shilin et al.}

\begin{abstract}

Sociocultural norms serve as guiding principles for personal conduct in social interactions, emphasizing respect, cooperation, and appropriate behavior, which is able to benefit tasks including conversational information retrieval, contextual information retrieval and retrieval-enhanced machine learning. 
We propose a scalable approach for constructing a Sociocultural Norm (\scns) Base using Large Language Models (LLMs) for socially aware dialogues. We construct a comprehensive and publicly accessible Chinese Sociocultural NormBase (\cnormb). 
Our approach utilizes socially-aware dialogues, enriched with contextual frames, as the primary data source to constrain the generating process and reduce the hallucinations. 
This enables extracting of high-quality and nuanced natural-language norm statements, leveraging the pragmatic implications of utterances with respect to the situation. 
As real dialogue annotated with gold frames are not readily available, we propose using synthetic data. 
Our empirical results show: (i) the quality of the {\scns}s derived from synthetic data is comparable to that from real dialogues annotated with gold frames, and (ii) the quality of the {\scns}s  extracted from real data, annotated with either silver (predicted) or gold frames, surpasses that without the frame annotations.
We further show the effectiveness of the extracted {\scns}s in a RAG-based (Retrieval-Augmented Generation) model to reason about multiple  downstream dialogue tasks.

\end{abstract}


\received{01 February 2024}
\received[revised]{01 June 2024}
\received[accepted]{5 September 2024}

\maketitle

\section{Introduction}

Sociocultural norms (\scns)s greatly influence the way people behave and the structures of society. Good knowledge of these norms can help better understand people's beliefs, attitudes, and actions. A better understanding of people is able to benefit several very recent and promising information retrieval and information generation related tasks including conversational information retrieval/seeking \cite{DBLP:series/irs/GaoXBC23,DBLP:conf/sigir/0001FORRTZ22,DBLP:conf/sigir/DeldjooTZ21}, contextual information retrieval \cite{DBLP:conf/ecir/HaiGFNPS23,DBLP:conf/sigir/SeylerCD18} and retrieval-enhanced machine learning \cite{DBLP:conf/sigir/BenderskyC0Z23, DBLP:conf/sigir/Zamani00MB22, 10.1145/3581783.3610949,10.1145/3581783.3612088}, commonsense-aware vision-language learning~\cite{10.1145/3539618.3591716,10447846,10.1145/3383184,10.1145/3648368}, 
semantic-aware vision-language discovery~\cite{10.1145/3522714,10.1145/3458281,10.1145/3635153}, 
especially in a cross-cultural setting. Additionally, a deeper understanding of sociocultural norms can also help alleviate the hallucinations associated with large language models (LLMs)~\cite{agrawal2023can,shi-etal-2023-hallucination,sun2023think}.

Sociocultural norms are often strongly associated with other relevant social factors~\cite{blass2015implementing}, such as formality, social relation and social distance. We refer to the combination of all social factors as \textbf{frames}. Different social factors, such as formality, often lead to different acceptable or unacceptable behaviors ~\cite{hovy2021importance}. 
For instance, given a context \textit{``Schedule a meeting for me''} in an informal setting, this can be done by \textit{``Text to a colleague or friend''}. 
However, such behavior is unacceptable for the formal setting. And an acceptable behavior should like \textit{``Request to a professional scheduler or secretary.''}. 

There have been recent attempts to establish sociocultural norm banks, in the form of rule-of-thumb natural language statements \cite{social101}, to enable their computational understanding. 
\normb\cite{ziems-etal-2023-normbank} has created situations by forming \textit{sociocultural frames} through  crowd-sourcing. 
However, \normb is in English, and labeled by English-speaking Mechanical Turk annotators who are located in the United States. Thus, {\scns}s in \normb may not be appropriate to the Chinese culture.
Besides, the manual construction of a norm bank is time-consuming and costly. 
\cite{normsage2022} has proposed an automated approach, \norms, to construct {\scns}s  by prompting~\cite{fei-etal-2023-reasoning} instruction-following LLMs. 
However, their lack of sociocultural dialogue frames leads to missing \scns statements related to pragmatic implications of utterances with respective to the social and interactional context \cite{LogicandConversation,holmes2017,zhou-etal-2023-cobra,zhan2023social}. 
Furthermore, their work is based on real dialogue data, leaving it unclear what can be done when such data does not exist. Moreover, we have experimentally found that, generated from real dialogue data which are usually ambiguous and noisy without frame as a constraint, the norms can be hallucinated. To the best of our knowledge, there is no frame-based Chinese norm base for studying Chinese sociocultural norms.

To close the research gap, we propose a scalable approach for \scns bank construction using LLMs for socially aware dialogues. Then we construct the first frame-based \textbf{C}hinese sociocultural \textbf{N}orm \textbf{B}ase, \textbf{\cnormb}, 
for investigating Chinese sociocultural norms.
Table~\ref{table:motivation} contextualises our work.
The norm base is grounded in the dialogues' contexts through sociocultural frames, enabling pragmatic reasoning related to the situational context and reducing the hallucinations. 
By applying these frames, we can embed domain knowledge into the model, 
significantly improving the control over social norm extraction.
To lessen the dependence on real dialogue data, we suggest employing synthetically generated dialogue data. This strategy is particularly useful when the required real data is unavailable. We also experimentally demonstrate the effectiveness of using synthetic data. For the annotation and extraction of SCNs, we utilize LLMs instead of human effort, offering a scalable solution for the construction of the \scns bank.
Once {\scns}s are extracted, we evaluate them both intrinsically and extrinsically. For the former, we use both automated (using GPT4) and human evaluation to assess the quality of the norm statements, and for the latter we employ {\scns}s in a fine-tuned Multilingual-BERT and a Retrieval-Augmented Generation (RAG)-based~\cite{rag2020,10.1145/3581783.3610949,10.1145/3581783.3612088} model which is a recent retrieval-augmented machine learning approach to reason about downstream tasks. 
%
To summarise, our contributions are,
\begin{itemize}[leftmargin=10pt] 
    \item We propose a novel pipeline that can automatically generate sociocultural norm (\scns) which leverages the frames to encode domain knowledge during the extraction process, such an approach significantly improves the control over the extraction of social norms and reduce the hallucinations during the generating process, leading to the production of high-quality \scns statements.
    \item We have constructed the Frame-based Chinese Socioculture Norm Base (\cnormb) 
    and make it available to the community.
    To the best of our knowledge, it is the first comprehensive social frame norm base focused on Chinese culture. For more information about the dataset, please refer to \href{https://github.com/SLQu/ChineseNormBase}{https://github.com/SLQu/ChineseNormBase}
    \item Our empirical analysis shows that (i) the quality of the {\scns}s extracted from synthetic data is on par with that from real dialogue data with gold (manually annotated) frames, and (ii) the quality of \scns bank extracted from real data annotated with silver (predicted) or gold frame is better than that without the frames.  
    \item We conduct comprehensive experiments to analyze and demonstrate the benefits of \cnormb in down-stream tasks using both fine-tuned Multilingual-BERT and retrieval-augmented generation models, including norm violation detection, and the prediction of the conversation frame elements, e.g. formality and social relation. 
\end{itemize}

\section{Related Work}


\paragraph{Commonsense Knowledge Bases (CSKB)} 
They include systematically organized information concerning everyday experiences\cite{speer2017conceptnet,elsahar2018t}, spanning extensive taxonomic relationships\cite{liu2004conceptnet}, logical interconnections~\cite{zhang2018record,elsahar2018t}, and fundamental principles of causality and physical mechanics~\cite{talmor2018commonsenseqa,bisk2020piqa}. Since the proposal of Cyc~\cite{lenat1995cyc}, a growing number of large-scale human-annotated  CSKBs~\cite{liu2004conceptnet,speer2017conceptnet,social101,bisk2020piqa,hwang2021comet,mostafazadeh2020glucose,ilievski2021cskg} have been developed. For instance, ConceptNet~\cite{speer2017conceptnet}, a large commonsense knowledge graph, embodies the traditional knowledge graph format comprising triples of head entities, relations, and tail entities. ATOMIC~\cite{sap2019atomic} contains nine social interaction relations and approximately 880,000 annotated triples. ATOMIC2020~\cite{hwang2021comet} integrates the relations of ConceptNet along with several novel relations, thereby establishing a more extensive CSKB composed of 16 event-related relations. Another notable CSKB is GLUCOSE~\cite{mostafazadeh2020glucose} which is constructed from text culled from ROC Stories~\cite{schwartz2017effect} and specifies ten commonsense dimensions to comprehensively study the causes and effects originating from a base event.


\begin{table}[t]
\centering
\setlength{\tabcolsep}{3pt}
    \begin{tabular}{c|cccc} \hline
      {Methods} & {\small Frame}  & {\small Chinese} & {\small Synthetic-Dial.} & {\small No Human} \\ \hline  
     \small \normb & \ding{52} & \ding{55} & \ding{55} & \ding{55} \\ 
     \small \norms  & \ding{55} & \ding{52}& \ding{55} & \ding{52} \\ \hline 
     \small \cnormb (ours)   & \ding{52} & \ding{52}& \ding{52} & \ding{52} \\ \hline 
    \end{tabular}%
    \caption{Categorization of prior work on sociocultural norm (\scns) statements extraction. ``Frame'' encompasses the socio-cultural context, e.g. the topic and social relation. ``Chinese'' refers the culture of norms. ``Synthetic-Dial.'' refers to using synthetic dialogues for extracting \scns.   ``No Human'' denotes  no need for human for \scns  extraction (different from evaluation).}
    \label{table:motivation}
\end{table}

\paragraph{Sociocultural NormBase Construction} 
SOCIAL CHEMISTRY \cite{social101} is a large-scale corpus of social and moral norms, which was constructed by crowdsourcing to elicit descriptive norms from situations via open-text rules-of-thumb as the basic conceptual units. 
\cite{zhan2023social} developed a dialogue corpus named \sdial, which is annotated with a variety of social factors to facilitate the study of Chinese social norms within conversations. Different in focus from ours, \sdial is a corpus of dialogues rather than a database of norms.
Additional pivotal work includes the study by \cite{ziems-etal-2023-normbank}. They developed a hierarchical taxonomy of constraints, named the Situational Constraints for Social Expectations, Norms, and Etiquette (SCENE). Following this, they engaged humans to annotate the rich SCENE taxonomy.
Our work primarily deviates from \normb as we introduce an automated pipeline for extracting socio-cultural norms, whereas \normb depends on human annotation.
Furthermore, another study \cite{normsage2022} proposed the \norms framework, which was designed to tackle task of uncovering norms grounded in conversation. This framework is built upon LLM prompting and self-verification, sourcing its discussions from authentic settings such as negotiations, informal chats, and documentaries. Our work differentiates itself from \norms by using social-cultural frame and synthetic-dialogues  for norm extraction, as shown in Table~\ref{table:motivation}.

\begin{figure*}[t]
  \centering
  \includegraphics[width=0.95\textwidth]{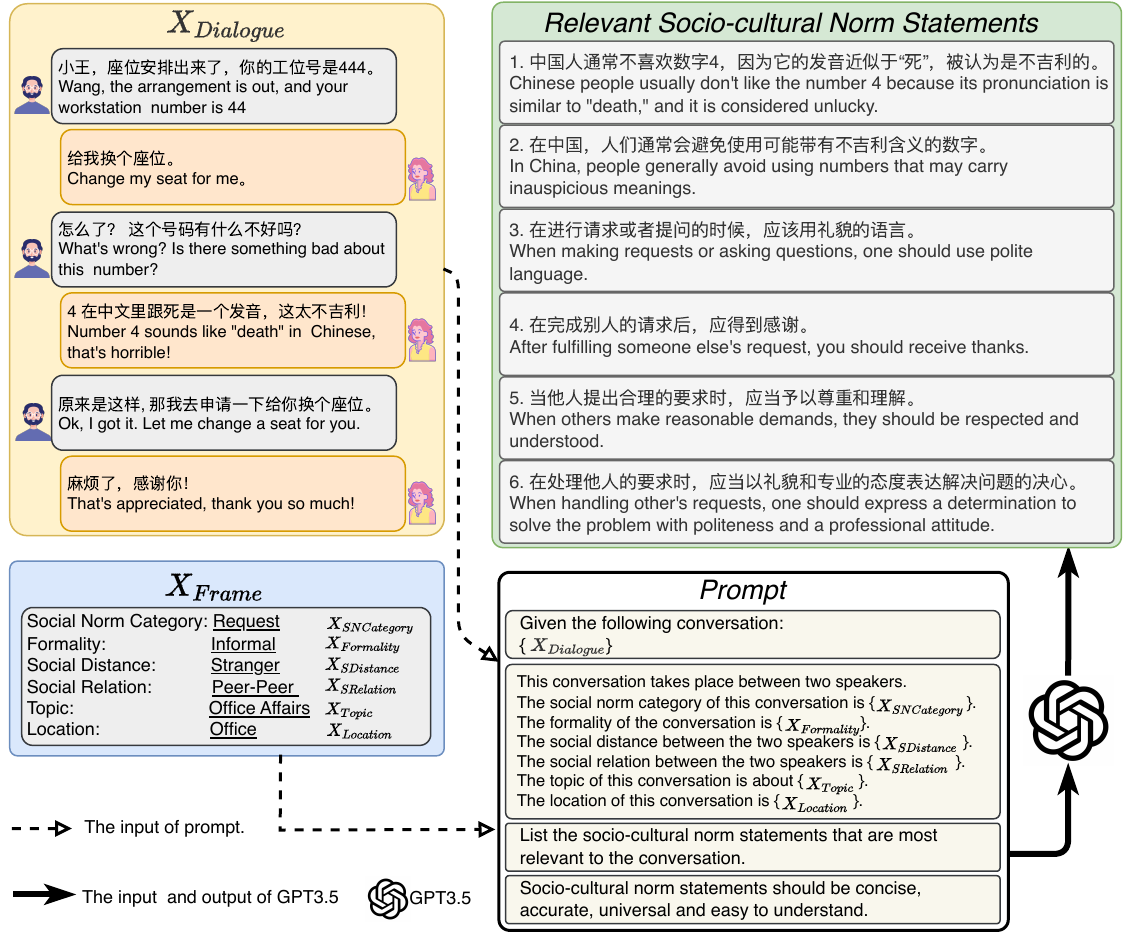}
  \caption{A demonstration of automatic socioculture norms extraction via LLM (GPT3.5) by giving a dialogue with frame information. }
  \label{fig:norm_extraction}
\end{figure*}

\section{Scalable Situated Norm Extraction}

We define the problem of extracting sociocultural norms {(\scns)}s from dialogues through their sociocultural frames. Given a dialogue $X_{dialogue}$ and its related sociocultural frame $X_{Frame}$, we aim to extract the relevant \scns natural-language statements ${X_{SCN}}_1,$ $...,$ ${X_{SCN}}_n$. 

We propose  a pipeline (detailed in \S \ref{sec:pipeline}) based on LLMs ~\cite{normsage2022,sap-etal-2020-social2,singhal2023large}. %
Our approach is innovative in several key aspects. 
By integrating a sociocultural frame into the pipeline, we enhance the  extraction of \scns statements due to leveraging the pragmatic implication of utterances. 
%
Furthermore, our pipeline is versatile, applicable not only to real dialogues but also effective in extracting high-quality \scns statements from synthetic dialogues, as analyzed in the experimental section~\ref{sec:Intrinsic Norm Discovery Evaluation}.

\label{subsec:SocioculturalDialogue_Frames}
\subsection{Sociocultural Dialogue Frames}
Context plays a pivotal role in understanding social interactions \cite{LogicandConversation,holmes2017}. 
We have adopted the frame elements from \sdial~\cite{zhan2023social}, which comprises six key social factors: Norm Category, Formality, Social Distance, Social Relation, Topic, and Location. 

Same as \sdial, in our work, Norm Category can take values from greetings, requests, apologies, persuasion, and criticism. Social Distance encompasses five distinct values: family, friends, romantic partners, working relationships, and strangers. Social Relation covers the following cases:  peer-to-peer, elder-junior, chief-subordinate, mentor-mentee, commander-soldier, student-professor, customer-server, and partner-partner. Formality is dichotomous, characterized as either formal or informal. Location spans various settings, including open areas, online platforms, homes, police stations, restaurants, stores, hotels, and refugee camps. Finally, Topic covers a wide array of subjects, such as sales, everyday life trivialities, office affairs, school life, culinary topics, farming, poverty assistance, police corruption, counter-terrorism, and cases of child disappearance.

As the real dialogue data annotated with gold frames may not be readily available, we propose to generate synthetic dialogues based on the content of the frames. Previous work has shown that LLMs can generate dialogue data with reasonable quality when prompted with key elements about the dialogue \cite{zhan2023social,li-etal-2023-normdial}. So we propose prompting LLMs with our sociocultural frames to generate corresponding synthetic dialogues (with the gold frame). This has the additional benefit that it can cover a wide range of situations via changig the frame. We show in our experiments that the quality of {\scns}s extracted from such synthetic data is on-par with real dialogues. We further show that when real dialogues without gold frames are available, a viable approach is to their predict silver frames, which then results in the extraction of reasonably good quality {\scns}s.

\subsection{Norm  Extraction Pipeline}
\label{sec:pipeline}

To extract a set of \scns statements from sociocultural frame based dialogues, we introduce the \prompt operator. The operator functions by executing four distinct parts:
\begin{itemize}
    \item A template header describing the nature of the dialogue data (i.e., conversation), followed by a fill-in slot $X_{dialogue}$ for the actual dialogue;
    \item The body of the template outlining the nature of the social factors (frame) of the dialogue. The sociocultural  factors, denoted by  $X_{Frame}$, include $X_{SNCategory}$, $X_{Formality}$, $X_{SDistance}$, $X_{SRelation}$, $X_{Topic}$, and $X_{Location}$.
    \item A directed question describing the task of \scns extraction.
    \item A verbalized constraint to ensure the format of the generated content (norm statements) is more unified and controllable.
\end{itemize}
Given a dialogue and its frame, we apply the \prompt and feed its output to the LLM to produce the  \scns statements.

\subsection{Quality Control}
Following the standard process for norm extraction, the last step should involve ensuring the quality of \scns statements. \schem  and \normb have utilized crowd-sourcing and manual verification of the \scns statements. \norms has introduced another round of prompting LLM to assure accuracy and relevance after generating the norms with LLM.  
Similarly, our approach also employs LLM to assess the accuracy and relevance of \scns statements in the context of the specified dialogue and its sociocultural framework.

\section{Evaluation}

We conduct two experiments to validate the quality and utility of our framework.
The first experiment, \textbf{Intrinsic Norm Discovery Evaluation}, examines the quality of the constructed \scns knowledge base.
The second experiment, \textbf{Extrinsic Evaluation on Downstream tasks}, seeks to demonstrate the applicability of \scns statements in an array of downstream tasks, such as detecting norm adherences and violations in dialogues and predicting social factors of dialogue. For the first one, the experiments are done based on both fine-tuned multilingual BERT model. For the latter one, the experiments are based on large language models (LLMs) equipped with retrieval-augmented generation (RAG). 

\paragraph{Dialogue Data, Frames, and \scns Statements}
We choose \sdial dataset as source of sociocultural frame based dialogues. \sdial comprises a total of 6,433 instances of multi-round dialogues (synthetic dialogues and real dialogues).
We extracted relevant \scns statements based on the dialogues of \sdial following the methodology illustrated in Figure 1. 
Each dialogue underwent this extraction process twice, extracting 140,669 \scns statements.

\paragraph{Implementation Considerations}
We use ChatGPT (GPT-3.5-turbo) as the underlying LLM in the \scns extraction framework of Figure~\ref{fig:norm_extraction}.    
For each dialogue, we set the maximum number of \scns statements that can be extracted to be $2\times$ the number of utterances in the dialogue. Dialogues contain varying numbers of utterances, making it difficult to define a universal upper limit for the number of sociocultural norms for all dialogues. However, it is easier to define it as a multiple of the number of utterances.
For dialogues lacking a frame, ChatGPT is first employed to predict a silver frame. The potential values for sociocultural factors are outlined in Subsection~\ref{subsec:SocioculturalDialogue_Frames}. Our approach includes a \scns statement pool. When a new \scns statement is generated, its similarity with existing norms in the pool is assessed. If the similarity score is below a predefined threshold of 0.97 which is chosen based on empirical study, the statement is considered novel and added to the pool.

\begin{table}[t]
    \begin{tabular}{c|ccccc} \hline
    & {\small Relevance} & {\small WellForm} & {\small Correct} & {\small Insight }& {\small Relatable } \\ \hline  \hline 
{\small \scns statements}   & \multicolumn{5}{c}{ {\small Evaluation via GPT4}}\\ \hline    
     \norms & 4.056 & \textbf{4.048}& 4.102& 3.729& 3.525\\
     \cnormb &\textbf{4.365}& 3.985& \textbf{4.269}& \textbf{3.988}& \textbf{3.998} \\ \hline \hline 
{\small \scns statements}   & \multicolumn{5}{c}{ {\small Evaluation via Human}}\\ \hline    
     \norms & 3.54 & 3.10 & 3.84 & 3.30 & 3.46\\
     \cnormb & \textbf{3.96} & \textbf{3.74} & \textbf{4.06} & \textbf{3.84} & \textbf{3.66} \\    \hline  
    \end{tabular}%
    \caption{Intrinsic evaluation of \scns statements extracted from \textit{synthetic dialogues} generated by ChatGPT.  The evaluations are based on Likert scores (1-5), judged by GPT4 and human as evaluators.}
    \label{table:likert_scale_syndata}
\end{table}

%

\begin{table}[t]
    \begin{tabular}{c|ccccc} \hline
    & {\small Relevance} & {\small WellForm} & {\small Correct} & {\small Insight }& {\small Relatable } \\ \hline  \hline 
{\small \scns statements}   & \multicolumn{5}{c}{ {\small Evaluation via GPT4}}\\ \hline    
     {\small \norms} & 4.023         & \textbf{4.041}        & 4.113         & 3.627         & 3.537\\
     {\small $\cnormb_{Silver}$} & 4.280         & 3.960         & \textbf{4.440}         & 3.962         & 3.983\\
     {\small $\cnormb_{Gold}$} & \textbf{4.345}& 3.975& 4.415& \textbf{3.964}& \textbf{4.092}\\ \hline \hline 
{\small \scns statements}   & \multicolumn{5}{c}{ {\small Evaluation via Human}}\\ \hline    
     {\small \norms} & 3.58 & 3.70 & 3.92 & 3.36 & 3.5 \\
     {\small $\cnormb_{Silver}$} & 3.94 & \textbf{3.76} & 4.10 & 3.64 & 3.66 \\
     {\small $\cnormb_{Gold}$} & \textbf{3.98} & 3.64 & \textbf{4.32} & \textbf{3.72} & \textbf{3.74} \\  \hline    
    \end{tabular}%

\caption{Intrinsic evaluation of \scns statements extracted from \textit{real dialogues} written by human.  The evaluations are based on Likert scores (1-5),  judged by GPT4 and human as evaluators.}
    \label{table:likert_scale_realdata}
\end{table}

\subsection{Intrinsic Norm Discovery Evaluation}
\label{sec:Intrinsic Norm Discovery Evaluation}
\subsubsection{Setting \& Baselines}
Similar to \norms~\cite{normsage2022}, our evaluation of norm extracted from dialogues utilizes a 1-5 Likert scale, where 1 means \enquote{awful} and 5 denotes \enquote{excellent}. We employ five evaluation criteria in our evaluation: \textit{Relevance}, \textit{Well-Formedness}, \textit{Correctness}, \textit{Insightfulness}, and \textit{Relatableness}. 
We randomly sample 100 synthetic dialogues and 100 real dialogues from \sdial, and follow our framework to generate relevant \scns. 

We use \norms~\cite{normsage2022} as our baseline, which is also a norm extraction pipeline but lacks the notion of frames in comparison to $\cnormb$. We implement \norms on the same dialogue dataset to extract norms based on its pipeline\footnote{The \norms dataset is not released when we do this work.}. We do this because there is no norm dataset released by \norms, and doing this can guarantee a fair comparison, which only evaluates the pipelines. We also study a scenario where the frame is not available, referred to as $\cnormb_{Silver}$. $Silver$ indicates the predicted frame by LLM based on the given dialogue, while $\cnormb_{Gold}$ represents real dialogues with real frames ($\cnormb_{Gold}$ is a subset of \cnormb). 
 We assess the extracted \scns statements using GPT4 and human evaluators using the five criteria mentioned earlier. 
We hire university students from China for the human evaluation part. To ensure fairness in their evaluations, we do not disclose where these \scns statements come from.
 The overall results can be found in Tables~\ref{table:likert_scale_syndata} and~\ref{table:likert_scale_realdata}.

\subsubsection{Result}
Table~\ref{table:likert_scale_syndata} demonstrates that on synthetic dialogues, \cnormb consistently outperforms \norms across 4 (out of 5) dimensions. \norms's primary limitation lies in its failure to consider the social frame information within dialogues, including aspects such as norm category, social distance, social relation, topic, location, and formality. These details are crucial for \scns statements, and our framework utilizes them as sources. As a result, our scores for \textit{Relevance}, \textit{Insightfulness} and \textit{Relatableness} show significant improvements in both real and synthetic dialogues. However, for the remaining two dimensions, \textit{Well-Formedness} and \textit{Correctness}, there isn't a significant difference between the two approaches, as frame information doesn't substantially impact \scns statement generation from these perspectives.
Table~\ref{table:likert_scale_realdata}, which focuses on real dialogues, validates that both $\cnormb_{Silver}$ and $\cnormb_{Gold}$ outperform \norms when leveraging real dialogues. This further underscores the importance of incorporating social frame information. Notably, $\cnormb_{Silver}$ achieves competitive results compared to $\cnormb_{Gold}$, indicating that even when the ground-truth frame is not available, the predicted frame using language models is still useful for norm extraction.

\begin{table}[htbp]
\centering
\caption{Norm adherence and violation label prediction on \ndial}
\label{tab:normdial}
\begin{tabular}{@{}l|cccc@{}}
\toprule
& \textbf{Data} & \textbf{Precision} & \textbf{Recall} & \textbf{F1-Score} \\ 
\hline \hline 
\multirow{4}{*}{\rotatebox[origin=c]{90}{Adherence}}& Multilingual-BERT & 60.22\% & 48.48\% & 0.5372 \\
& NormBank-BERT & 62.50\% & 60.27\% & 0.6136 \\
& NormSage-BERT & 63.23\% & 54.14\% & 0.5833 \\
& ChineseNormBase-BERT & \textbf{64.58}\% & \textbf{69.51}\% & \textbf{0.6695} \\ \midrule
\multirow{4}{*}{\rotatebox[origin=c]{90}{Violation}}  & Multilingual-BERT & 62.62\% & 71.53\% & 0.6678 \\
& NormBank-BERT & 65.62\% & 75.81\% & 0.7035 \\
& NormSage-BERT & 67.09\% & 76.09\% & 0.7131 \\
& ChineseNormBase-BERT & \textbf{68.71\%} & \textbf{76.71\%} & \textbf{0.7249} \\ \bottomrule
\end{tabular}
\end{table}


%
%
\begin{table*}[t]
\caption{
    Experimental results of exploring the impact of  {\scns}s on dialogue social factors. The relevant  {\scns}s are placed in the prompt of the underlying LLM reasoner together with the test case for the prediction of the target, amounting to a RAG-based approach for reasoning based on our \cnormb.  Please note that, according to documents of sklearn, \href{https://scikit-learn.org/stable/modules/generated/sklearn.metrics.f1_score.html}{when using `macro' to account for label imbalance, it can result in an F-score that is not between precision and recall.}
}
\tiny
\begin{tabular}{c|c|c|c|c|c|c}
\hline 
 & {Norm} & Formality & Location & Distance & Relation & Topic \\  
 &{\small Pre/Rec/F1} & {\small Pre/Rec/F1} & {\small Pre/Rec/F1} & {\small Pre/Rec/F1} & {\small Pre/Rec/F1} & {\small Pre/Rec/F1}     \\  \hline \hline 
{\small\# of {\scns}s} & \multicolumn{6}{c}{ ChatGPT as the LLM Reasoner}\\ \hline 
0  & .4639/.5931/.1944 & .5179/.7500/.4290 & .6885/.6201/.4922& .5440/.5770/.4239 & .4933/.5353/.3688 & .6253/.6396/.4573\\ 
1  & \textbf{.4847}/\textbf{.6077}/\textbf{.2214} & .5479/.7815/.4800 &  \textbf{.7437}/\textbf{.6277}/\textbf{.5191}& .5567/.6030/.4440 & .5151/.5480/.3996 & .6348/.7395/.5430 \\
all & .4782/.6075/.2161 & \textbf{.5909}/\textbf{.8037}/\textbf{.5550} & .7119/.5762/.4844 & \textbf{.5634}/\textbf{.6090}/\textbf{.4616}& \textbf{.5317}/\textbf{.5813}/\textbf{.4051} & \textbf{.6567}/\textbf{.7676}/\textbf{.5607}\\  \hline

{\small\# of  {\scns}s} & \multicolumn{6}{c}{GPT4 as the LLM Reasoner}\\ \hline
0   & .5603/.6180/.2611 & .6080/.6340/.3846& .6115/\textbf{.7668}/.5461 & .6432/.6854/.5163 & .5094/.6148/.4590 & .6913/.8000/.5955 \\ 
1   & .5768/.6285/.2954 & \textbf{.6331}/.6762/.4566& .6485/.7099/.5294& \textbf{.7378}/.6942/.6283 & .5128/\textbf{.6460}/\textbf{.4741} & \textbf{.7829}/.8167/\textbf{.7027} \\ 
all & \textbf{.5776}/\textbf{.6449}/\textbf{.2992} & .6156/\textbf{.7449}/\textbf{.4794}& .\textbf{6733}/.7453/\textbf{.5630} & .7155/\textbf{.7393}/\textbf{.6302}& \textbf{.5297}/.6364/.4725 & .7613/\textbf{.8171}/.6833  \\ \hline
\end{tabular}%

\label{table:violation}

\end{table*}

\begin{figure}[h]
  \centering
  \includegraphics[width=0.5\textwidth]{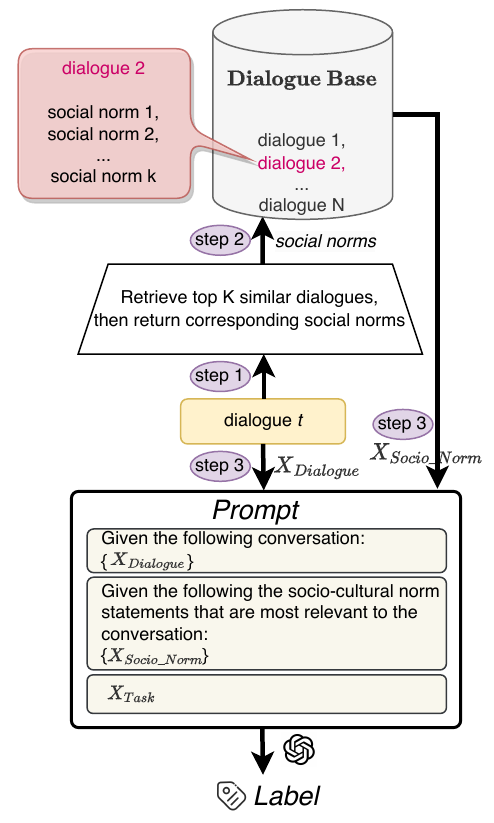}
  \caption{The framework of retrieving \scns statements for social factors prediction. Given target dialogue $t$, retrieve top $K$ similar dialogues from the data base to get corresponding \scns statements for relevant tasks}
  \label{fig:social_norm_retrieve}
\end{figure}

\subsection{Extrinsic Evaluation on Norm Adherence and Violation Detection}
We use the task of detecting dialogue norm adherence and violation to validate the importance of having a frame-based Chinese sociocultural norm base.

\subsubsection{Dataset}
\ndial~\cite{li-etal-2023-normdial}, labeled with social norm adherences and violations on a dialogue-turn, is dialogue dataset in Chinese and  American cultures. We select the Chinese dialogues in \ndial for this experiment. We divide the labeled dialogues from \ndial into a 6:2:2 ratio for training, validation, and testing purposes.

\subsubsection{Setting \& Baselines}
We fine-tune a pretrained Multilingual-BERT\footnote{https://github.com/google-research/bert/blob/master/multilingual.md} on three norm datasets: \normb~\cite{ziems-etal-2023-normbank}, \norms~\cite{normsage2022} (i.e., the norms generated by \norms using \sdial dataset), and \cnormb (i.e., the norms generated by \cnormb also using \sdial dataset) respectively, using masked language modeling.  During this fine-tuning process, we randomly mask 15\% of the words in each input sentence, and the model learn to predict these masked words. As a result, we learn and compare three models:  \normb-BERT, \norms-BERT, and \cnormb-BERT. For these three models, after fine-tuning them on the three norm datasets with masking scheme, we further fine-tune them on the training set of \ndial for the downstream task and then use them to predict on the test set. We also compare these three models with a Multilingual-BERT model which is directly fine-tuned on the training set of \ndial for the downstream task without fine-tuning on the norm datasets. 
Our evaluation criteria include precision, recall, and F1-Score.

\subsubsection{Result}
Table~\ref{tab:normdial} displays the results of norm adherence and norm violation prediction on \ndial. It is evident that \cnormb-BERT outperforms all the baselines in both norm adherence and norm violation prediction. While \norms-BERT also concentrates on Chinese culture, \cnormb-BERT gains an advantage by incorporating social frame information. \normb-BERT, despite having frame information, primarily focuses on American culture, whereas \cnormb-BERT is tailored for Chinese culture. Multilingual-BERT consistently performs the worst, mainly as it has not been pretrained or fine-tuned on any social norm datasets.

\subsection{Extrinsic Evaluation on Social Factors Prediction}
\label{subsec:downstream}
In this section, we evaluate \cnormb on the downstream task of social factors prediction. The foundation model we use here is Retrieval-Augmented Generation (RAG) based large language models (LLMs)~\cite{rag2020}. More specifically, given the target dialogue, top-k most similar dialogues are retrieved from a dialogue database. The target dialogue together with \scns statements of these top-k similar dialogues are input to the LLM for prediction. Here, we use \sdial as the dialogue database to assess if it can help with the downstream task using RAG-based models.
Note that this retrieval process is accomplished by embedding all dialogues using the pre-trained language model BERT~\cite{cui2020revisiting}.
Then, we calculate cosine similarity to find the top-k similar dialogues of the target dialogue.  
The overview of the process is illustrated in Figure~\ref{fig:social_norm_retrieve}.

\subsubsection{Setting \& Baselines}
We use \sdial~\cite{zhan2023social} annotated with rich social factors as the evaluation dataset for the downstream task. 
The predicted social factors are norm category, formality, location, social distance, social relation, and topic. We use ChatGPT and GPT4 as the LLM reasoner, and use macro-precision, macro-recall, and macro-F1 as metrics with $k = 5$.
In Table~\ref{table:violation}, `\textbf{0}' indicates that only the original dialogue (no \scns statement at all) is used for prediction. `\textbf{all}' means that all the retrieved \scns statements are used for prediction. And `\textbf{1}' means using a randomly selected \scns statement from all the retrieved \scns statements for prediction.

\subsubsection{Result}
As we can see from Table~\ref{table:violation}, for the prediction of norm category, formality, social distance, social relation, and topic, both ChatGPT and GPT4 show a stronger performance with \scns statements than without them. This result suggests that \scns statements do facilitate dialogue-oriented downstream tasks. Moreover, we can observe that in most cases, \textbf{all} \scns statements induce better performance than \textbf{1} \scns statement. This indicates that the more relevant \scns statements are provided, the greater the improvements to predicting social factors in dialogues. As for location prediction, especially recall, the results without \scns statements are better than those given \scns statement. We posit this happens because \scns statements are less closely related to where a dialogue occurs.
Providing such \scns adds noisy information, and can hurt location prediction. 

In some instances, we noticed that the macro-F1 score is lower than the macro-precision and macro-recall scores. This can be attributed to the fact that predicting social factors in \sdial data is a multi-class classification problem with significant variations in sample sizes among categories. 
For instance, in the case of social distance, there are 5 categories, and the `working' category makes up 59.30\% of them. Similarly, for social relation, there are 7 categories, but `mentor-mentee' represents just 0.41\%. 
As a result, models may struggle to achieve high macro-precision and macro-recall in specific categories, consequently affecting the overall multi-class macro-F1 score. 
And according to documents of sklearn, \href{https://scikit-learn.org/stable/modules/generated/sklearn.metrics.f1_score.html}{when using `macro' to account for label imbalance, it can result in an F-score that is not between precision and recall.}
\footnote{\href{https://scikit-learn.org/stable/modules/generated/sklearn.metrics.f1_score.html}{https://scikit-learn.org/stable/modules/generated/sklearn.metrics.f1\_score.html}} This observation is consistent with findings in \sdial~\cite{zhan2023social}.

\section{\scns Statement Analysis}
In this section, we will demonstrate that synthetic data, which includes dialogues and social frames, maintains sufficient quality for extracting \scns statements when compared to real-world dialogues and social factor frames. Additionally, we will present an overview of the distribution of \scns statements across various social factors. Finally, a case study will be provided to show the benefit of frames.

\subsection{\scns from Synthetic Data}
\label{sec:from Synthetic Data}
\subsubsection{Design}
In this section we study the quality of the \scns generated from $Real$ and $Synthetic$ dialogues. We also delineate the effect of the ground-truth social frames and the predicted social frames, referred to as $Gold$ and $Silver$, respectively. For our analysis, we randomly selected 100 $Real$ dialogues and 100 $Synthetic$ dialogues, each paired with their respective $Gold$ frames. We ensured that both sets of dialogues shared the same frame combination.
Following the framework in Figure~\ref{fig:norm_extraction}, we extracted \scns statements from both $Real + Gold$ and $Synthetic + Gold$. Then, we applied the same extraction pipeline on 100 $Real$ dialogues with $Gold$ frames and 100 $Real$  dialogues with $Silver$ frames, represented as $Real + Gold$ and $Real + Silver$, respectively. 

\subsubsection{Evaluation Criteria}
In our evaluation, we assess the overlap between two sets of statements using precision, recall, and F1 score. Let's denote the two statement sets as $A$ and $B$, where set $A$ is ground-truth. 
We calculate the Cosine similarity of the BERT embeddings between statements in $A$ and $B$. If the similarity between a statement from $A$ and a statement from $B$ exceeds a pre-defined threshold (which we empirically set at 0.97), we consider the two statements to be the same. i.e., they belong to the intersetion of $A$ and $B$. 
Based on the above procedure, Precision, Recall, and F1 are defined as follows:
$$Precision =\frac{\left| A \cap B \right|}{\left| A \right|}$$
$$Recall =\frac{\left| A \cap B \right|}{\left| B \right|}$$
$$F1 =\frac{2 \times Precision \times Recall}{Precision + Recall}$$

\begin{table}[ht]
    \caption{\scns statements overlap analysis. `Real vs Synthetic' compares \scns statements derived from real dialogues and synthetic dialogues, both incorporating gold frames. `Real gold vs pred' refers to the comparison of \scns statements from real dialogues using gold frames against those using predicted frames.}
    \begin{tabular}{c|cccc} \hline
    Pair                & \small\# of \scns & Pre/Rec/F1      \\ \midrule
    $Real$ + $Gold$ $v.s.$ $Synthetic$ + $Gold$  & 486/554 & .928/ .959/ .942 \\
    $Real$ + $Gold$ $v.s.$ $Real$ + $Silver$   & 486/484 & \textbf{.938}/ \textbf{.961}/ \textbf{.949} \\ \bottomrule
    \end{tabular}
    \label{table:overlap}
\end{table}

\subsubsection{Result}

As shown in Table~\ref{table:overlap}, for the 100 real dialogues with gold frame, 100 synthetic dialogues with gold frame, and 100 real dialogue with silver frame, there are 486, 554, and 484  extracted \scns statements. The precision, recall, and F1 scores of the \scns statemens on  $Real$ + $Gold$ vs $Synthetic$ + $Gold$ and $Real$ + $Gold$ vs $Real$ + $Silver$ both exceed 0.9. This result demonstrates the high quality of the synthetic dialogues and predicted frames: that they induce highly similar \scns to those induced from real dialogues and gold frames. Moreover, the experiment on $Real$ + $Gold$ $v.s.$ $Synthetic$ + $Gold$ show a high overlap in \scns despite the different content in the two dialogue datasets, underscoring the significance of frames in \scns statements extraction. Additionally, the higher overlap observed in the experiment $Real$ + $Gold$ $v.s.$ $Real$ + $Silver$ suggests that \scns statements are more influenced by the dialogue itself than by the frames.

\subsection{Distribution of \scns Statements on Sociocultural Factors}
\begin{figure}[h!]
  \centering
  \includegraphics[width=0.45\textwidth]{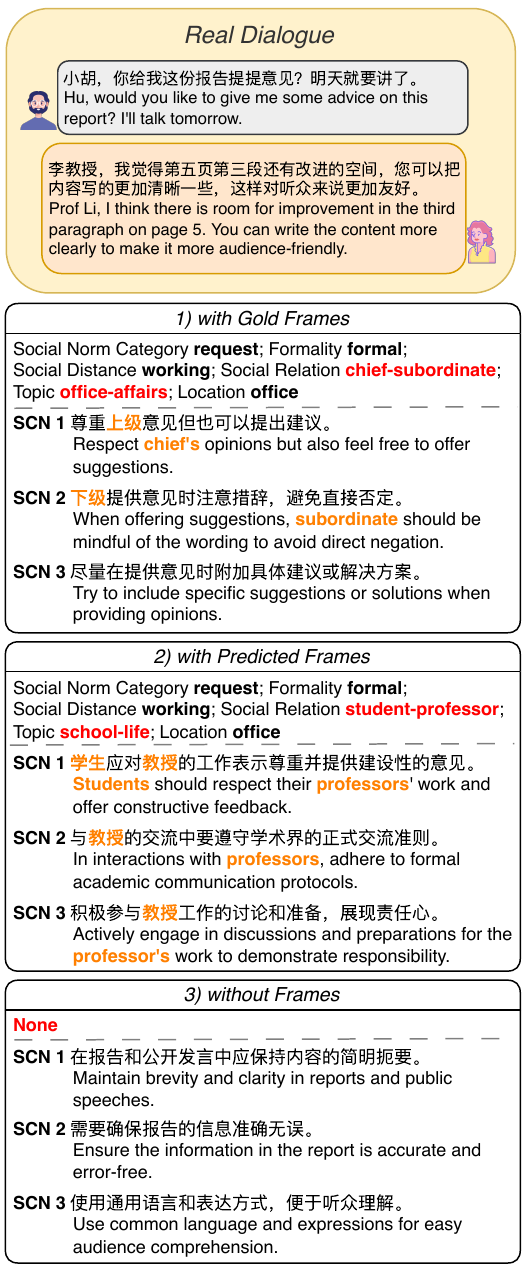}
  \caption{Sample \scns statements extracted from different datasets. 
  }
  \label{fig:scn_examples}
\end{figure}
\subsubsection{Setting}
Similar to Subsection~\ref{sec:from Synthetic Data}, we extract \scns statements from real dialogues with gold frames, real dialogues with silver frames, and synthetic dialogues with gold frames. Subsequently, GPT-4 was employed to classify the extracted \scns statements into different categories of a specified social factor.
\subsubsection{Result}

\begin{figure*}[t]
  \centering
  \includegraphics[width=0.95\textwidth]{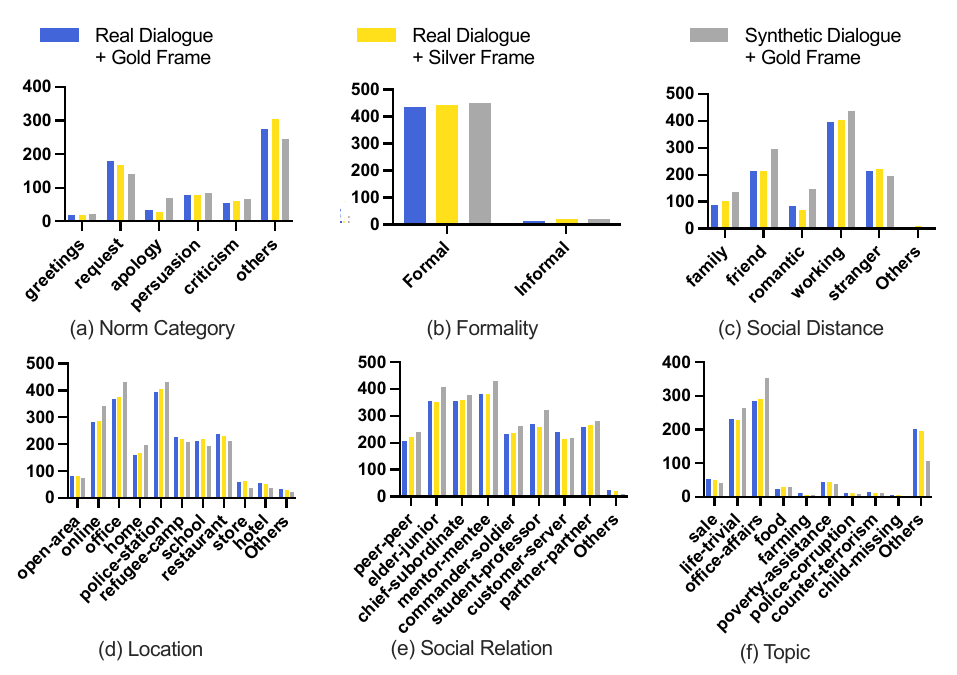}
  \caption{Distribution of \scns statements on different categories on six social factors. }
  \label{fig:main}
\end{figure*}

From Figure~\ref{fig:main}, in each sub-plot, the vertical axis quantifies the frequency of each category, while the horizontal axis represents categories with the three bars corresponding to different dialogue and frame combinations.

For social factor norm categories, the candidates can be greetings, requests, apologies, persuasion, criticism, and others. The resulting distribution of \scns statements from different data sources is depicted in Figure~\ref{fig:main}(a).
Our analysis reveals a non-uniform distribution of social norm categories across all datasets, a trend consistent in each dataset. This indicates the varied impact of different norm categories in daily life. Notably, the `others' category significantly surpasses others, suggesting the existence of numerous norm categories not encompassed by the predefined five categories. Additionally, the `request' and `persuasion' categories exhibit moderate frequencies, implying these social norms are more prevalent in everyday interactions.

Among the various social factors analyzed, it is observed that `Formal' is significantly more prevalent than `Informal' in `Formality'. In `Social Distance', the `Working' relationship is the most common. As for `Social Relation', the patterns of `Elder-Junior', `Chief-Subordinate', and `Mentor-Mentee' relationships are predominant, all indicative of superior-subordinate dynamics. The most frequent `Location' for these interactions is the `Office', and the common `Topic' revolves around `Office Affairs'. This analysis suggests that for the \sdial dataset, a significant portion of social norms pertains to interactions between superiors and subordinates within formal work environments.

\subsection{Case Study}
Figure~\ref{fig:scn_examples} presents some sample \scns statements generated from real dialogues under three different conditions: with Gold Frames, with Silver Frames, and without Frames.
Firstly, in the first sub-table with Gold Frames, the \scns statements encapsulate the ``chief-subordinate" relationship within the context of ``office affairs". Similarly, in the second sub-table with Silver Frames, the \scns statements encompass the ``student-professor" dynamic in ``school life".
However, in the third sub-table, where dialogues are presented without frames, all the \scns statements revolve solely around conducting reports. This approach overlooks the broader context of the real dialogues, resulting in a lack of relevance and depth in the extracted \scns statements.
These observations underscore the value of incorporating frames in \scns statements extraction. By doing so, we can effectively infuse domain knowledge into the process, transcending the limitations of relying solely on the dialogue content.

\section{Conclusion}
We propose a scalable approach for constructing a Sociocultural norm base using large language models (LLMs) for socially-aware dialogues. This norm base is rooted in the dialogues' contexts, enriched with sociocultural frames, thus enabling pragmatic reasoning relevant to the situational context. As real dialogue annotated with gold frame are not readily available and expensive to collect, we also show that it is possible to extract high-quality \scns statements from synthetically generated data. This is particularly encouraging for low-resource languages and cultures, to the extent to which they can be covered by LLMs.  
We further show the effectiveness of the extracted {\scns}s in a RAG-based model to reason about multiple downstream dialogue tasks. 
We believe \cnormb is a valuable resource for studying the Chinese culture, and our norm extraction pipeline a scalable process that can be applied for many languages and cultures.

\section*{Limitations}
Our framework is based on the implicit commonsense knowledge and reasoning in 
ChatGPT (GPT-3.5-turbo) from OpenAI\footnote{https://openai.com}, as the underlying LLM, for pragmatic reasoning of situated dialogues and extracting the norm statements. 
ChatGPT is trained on large amount of online data. However, its commonsense knowledge and reasoning can be biased, although  efforts has put into guardrails to address bias and increase the model safety. Furthermore, sociocultural norms can evolve and shift over the time, which requires the adaptation of the commonsense knowledge and reasoning of the underlying LLM with the evolving data.
Due to limitation of time and computational resources, we did not rigorously study the extent to which recent publicly accessible LLMs are capable of extracting high-quality norm statements, and leveraging pragmatic reasoning on situated dialogues. 

\section*{Ethical Statement}
We acknowledge that automated generation of sociocultural norm statements an be authoritative and normative \cite{normsage2022}. We  thus emphasise that they are not intended to be used for forming a norm system for any society. 
Furthermore their use in any deployed system needs to be done with care. It needs to involve human in the loop to inspect the validity of them before the deployment. As such, these norm statements are mainly for research purposes. 
Our approach amounts to an explainable verification framework to assess the sociocultural  knowledge and reasoning of LLMs. It can help to extract LLMs' norm rules (as described in our work), whose validity can then be assessed by human.
%


\begin{acks}
This material is based on research sponsored by DARPA under agreement number HR001122C0029. The U.S. Government is authorized to reproduce and distribute reprints for Governmental purposes notwithstanding any copyright notation thereon.
\end{acks}

\bibliographystyle{ACM-Reference-Format}
\bibliography{anthology,custom}


\end{document}